# A Framework to Enhance Generalization of Deep Metric Learning methods using General Discriminative Feature Learning and Class Adversarial Neural Networks

Karrar Al-Kaabi[1], Reza Monsefi*[2], Davood Zabihzadeh[3]

[1] Faculty of Veterinary Medicine, University of Kufa Al-Najaf, Kufa, IRAQ
[2] Computer Department, Engineering Faculty, Ferdowsi University of Mashhad, Mashhad, IRAN
[3] Computer Engineering Department, Hakim Sabzevari University, Sabzevar, IRAN

* Corresponding Author
     karrara.hussein@uokufa.edu.iq, monsefi@um.ac.ir, d.zabihzadeh@hsu.ac.ir

## Abstract

Metric learning algorithms aim to learn a distance function that brings the semantically similar data items together and keeps dissimilar ones at a distance. The traditional Mahalanobis distance learning is equivalent to find a linear projection. In contrast, Deep Metric Learning (DML) methods are proposed that automatically extract features from data and learn a non-linear transformation from input space to a semantically embedding space. Recently, many DML methods are proposed focused to enhance the discrimination power of the learned metric by providing novel sampling strategies or loss functions. This approach is very helpful when both the training and test examples are coming from the same set of categories. However, it is less effective in many applications of DML such as image retrieval and person-reidentification. Here, the DML should learn general semantic concepts from observed classes and employ them to rank or identify objects from unseen categories. Neglecting the generalization ability of the learned representation and just emphasizing to learn a more discriminative embedding on the observed classes may lead to the overfitting problem. To address this limitation, we propose a framework to enhance the generalization power of existing DML methods in a Zero-Shot Learning (ZSL) setting by general yet discriminative representation learning and employing a class adversarial neural network. To learn a more *general representation*, we propose to employ feature maps of intermediate layers in a deep neural network and enhance their discrimination power through an attention mechanism. Besides, a *class adversarial network* is utilized to enforce the deep model to seek class invariant features for the DML task. We evaluate our work on widely used machine vision datasets in a ZSL setting. Extensive experimental results confirm that our framework is indeed helpful to improve the generalization of existing DML methods and it consistently outperforms baseline DML algorithms on unseen classes.

## 1. Introduction

Distance measures have a major role in the success of many machine learning or pattern recognition algorithms. Standard distance or similarity measures such as Euclidean or cosine similarity often fail to capture the semantic concepts needed for a specific task. Thus, we need data dependent metrics that measure semantically similar pairs close together while evaluating dissimilar ones far apart.

Most research in metric learning is dedicated to Mahalanobis distance defined as:

$$d_M(x_i, x_j)^2 = (x_i - x_j)^T M(x_i - x_j) \tag{1}$$

where $x_i, x_j \in \mathbb{R}^d$ and $M \in \mathbb{R}^{d \times d}$ is a positive semi-definite (p.s.d) matrix. Since $M \succcurlyeq 0$, it can be factorized as $LL^T$, where $L \in \mathbb{R}^{d \times r}$ and $r = rank(M) \leq d$. Thus,

$$d_M(x_i, x_j)^2 = (x_i - x_j)^T M(x_i - x_j) = (x_i - x_j)^T LL^T (x_i - x_j) = \|L^T(x_i - x_j)\|_2^2 \tag{2}$$

The above equation indicates that Mahalanobis distance learning is equivalent to find a *linear* transformation matrix $L$. Despite its success, in many nonlinear datasets with complex class boundaries, a linear transformation is unable to extract discriminative features. Besides, Mahalanobis metric learning methods cannot process complex data such as images, text, and videos directly. Therefore, we need to first extract features from data and then apply a metric learning algorithm.

To overcome these issues, deep metric learning methods are proposed that jointly perform feature extraction and learning non-linear transformation. Hence, they obtain an optimized representation of the input data for the distance learning task. Specifically, DML[1] finds a non-linear function $f_w(x)$ parameterized by w that maps $x$ from the input space to a semantic embedding space. In this space, similar data items should be close together while dissimilar pairs should be kept at a distance.

A typical DML model has three components: 1) deep neural network model, 2) sampling algorithm, 3) loss function.

The function $f_w$ is implemented using a deep neural network such as CNN where the Softmax output layer is replaced by an FC[2] named embedding layer. The embedding consists of a weight matrix $L \in \mathbb{R}^{h \times d}$ that projects the output of the last hidden layer into the semantic embedding space.

To learn the parameters of the deep model (i.e., $w$), the network is often trained by pairwise or triplet constraints defined as follows:

- Pairwise: $S = \{(x_i, x_j) \mid x_i \text{ and } x_j \text{ are similar}\}$ (Similar set), and
$$D = \{(x_i, x_j) \mid x_i \text{ and } x_j \text{ are dissimilar}\} \text{ (Dissimilar set)}$$

---

[1] Deep Metric Learning

[2] Fully Connected

- Triplets: $T = \{(x_i, x_i^+, x_i^-) \mid x_i \text{ should be more similar to } x_i^+ \text{ than to } x_i^-\}$. Here, $x_i$, $x_i^+$, and $x_i^-$ are named anchor, positive, and negative data, respectively.

Many DML algorithms adopt the Siamese (Chopra, Hadsell et al. 2005) or the Triplet (Wang, Song et al. 2014, Hoffer and Ailon 2015) architecture shown in Figure 1. Also, other forms of constraints such as quadruplets (Ni, Liu et al. 2017), and n-pairs (Sohn 2016) are defined in recent research.

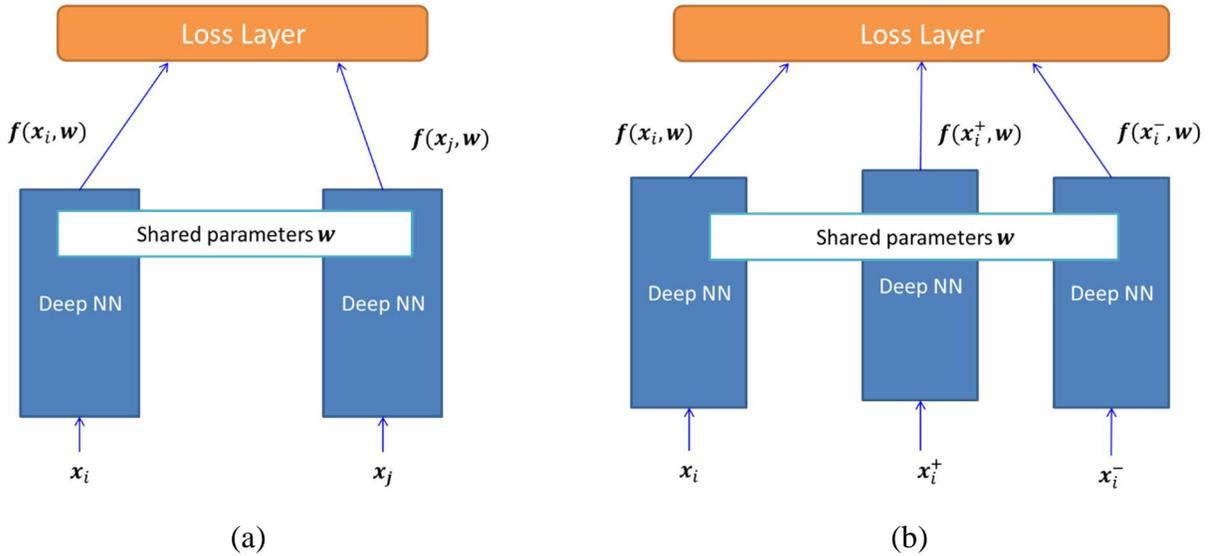

**Figure 1- (a) The Siamese architecture. (b) The Triplet architecture**

Many sampling strategies are developed to obtain informative constraints such as pairs, and triplets from the input minibatch of data passed to the network. Some seminal mechanisms include easy (Kaya and Bilge 2019), hard (Simo-Serra, Trulls et al. 2015), and semi-hard (Schroff, Kalenichenko et al. 2015) negative mining. The sampled constraints are forwarded to the loss layer that encourages the separation of positive and negative pairs. The loss gradient is then backpropagated through the network to adjust the deep network parameters.

A loss function has a key role in the training process of a deep model. Two popular DML loss functions are *contrastive,* and *triplet* utilized in the Siamese and Triplets networks, respectively. Besides, many other loss functions are developed to promote the performance of the given DML task.

Many DML methods emphasize learning a more discriminative embedding function by providing novel sampling strategies or loss functions. This approach yields a good result where training and testing data are collected from the same set of categories. However, this assumption is not correct in major applications of DML such as CBIR[1], person re-identification, ZSL[2], and FSL[3]. Here, the model should learn general informative concepts from seen classes and utilize them to rank or identify objects from unseen categories. Neglecting the generalization power of the semantic embedding and just seeking to learn more discriminative representations of seen classes is subject to the overfitting problem.

---

[1] Content-Based Information Retrieval

[2] Zero-Shot Learning

[3] Few-Shot Learning

To address this challenge, we propose a novel framework to enhance the generalization of DML methods in the ZSL setting using general yet discriminative feature learning and a class adversarial neural network.

In a typical deep neural network, the middle layers of the network extract general and small patterns of given data. The output of each layer is passed to the next layer to extract more discriminative and large patterns from the input. The last hidden layer of the network produces the most discriminative features, but it is dependent on the observed classes. It also concentrates only on specific regions from the input image needed to discriminate training classes. That increases the risk of neglecting other important regions for unseen categories. This problem is illustrated in Figure 2. The figure shows the generated Grad-CAMs of some images using fine-tuned BN-Inception neural network on the CUB200-2011 dataset. As seen, the deep model mainly focused on the head of birds to classify the images and omit some other discriminative regions that might be helpful to identify unseen objects.

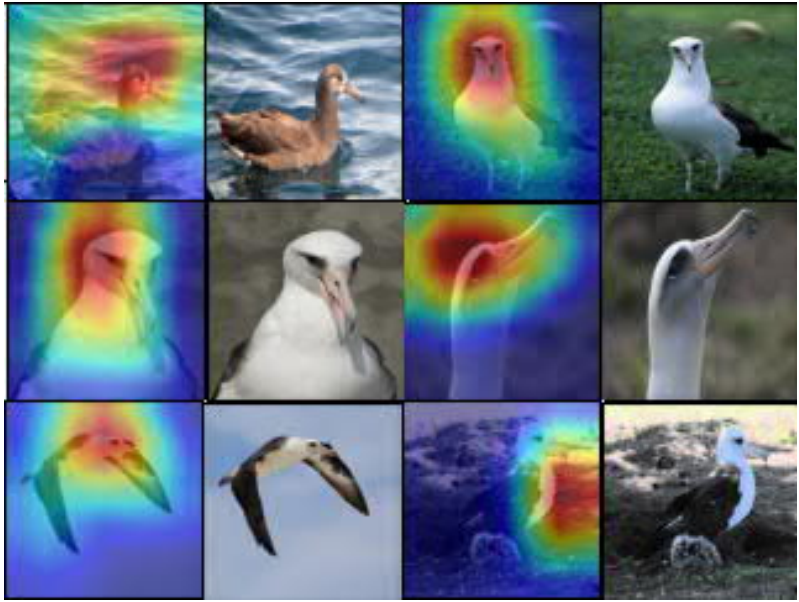

**Figure 2- Generated Grad-CAMs of some Birds using fine-tuned BN-Inception neural network on CUB200-2011 dataset.**

Let $u$ be the output of the last hidden layer in the network. Most existing DML approaches use only the feature vector $u$ to learn a discriminative embedding. Here, we propose to employ outputs of intermediate layers since they have a better generalization on unseen categories. Also, they cover most important regions from the input image. The problem is the low discrimination power of these feature maps. We handle this issue by attending these features to the discriminative $u$ feature vector. We weight the feature maps of the selected layers according to discrimination scores obtained by an attention mechanism and form the final feature vector by combining the weighted feature maps.

Besides, an adversarial network is employed that enforces the deep model seeking class invariant features for the DML task. The adversarial network increases the classification loss on seen classes during the semantic embedding learning process. This approach causes the deep network to explore more general discriminative information that is less dependent only on the available classes for the metric learning task. Therefore, it improves the generalization capability of the learned semantic embedding and prevents the overfitting on observed categories.

In summary, the major contributions of this paper are as follows:
  I. Learning general and discriminative representation by utilizing intermediate feature maps of the deep model and enhancing their discrimination power by attending them to the feature vector obtained from the last hidden layer.
  II. A general framework to enhance the generalization power of DML is proposed using a class adversarial neural network. The framework can be applied to many existing DML methods and improves their performance in a ZSL setting.
  III. We investigate several strategies to implement a class adversarial module and observed the best performance is obtained by confusing the classifier via adaptively setting the adversarial coefficient and label smoothing.
  IV. Extensive experiments are performed on popular image information retrieval datasets in a ZSL setting, and our approach improves the results of state-of-the-art DML methods on these datasets.

The rest of the paper is organized as follows. Section 2 reviews related deep semantic embedding learning methods. Section 3 presents our framework to enhance the generalization power of DML in a ZSL setting. Strategies to implement the framework are presented in Section 4. Experimental results and comparison with state-of-the-art methods are reported in Section 5. Finally, the conclusion of the work and recommendations for future studies is given in Section 6.

## 2. Related Work

(Chopra, Hadsell et al. 2005) developed based on a Siamese network is the first pioneer work in the deep metric learning domain. Here, the energy between two pairs is defined as:

$$E_W(x_1, x_2) = \|f_W(x_1) - f_W(x_2)\| \tag{3}$$

The aim is to learn the network weights to minimize the energy function for similar images and maximize it for dissimilar ones. To this end, the following loss function is proposed:

$$L(W, Y, X_1, X_2) = (1 - Y)L_G(E_W) + YL_I(E_W)$$
$$= (1 - Y)\frac{2}{Q}(E_W)^2 + Y2Q \exp(-\frac{2.77}{Q}E_W) \tag{4}$$

where binary variable Y indicates that the corresponding $(X_1, X_2)$ is similar or dissimilar pair:

$$Y = \begin{cases} 0, & for\ similar\ pair\ (X_1, X_2)\ (genuine\ pair) \\ 1, & otherwise \end{cases}$$

Loss functions defined on pairwise constraints such as (3) are named *Contrastive*. These functions are based on absolute distance values. However, in many applications, the *relative* distances between positive and negative pairs are more important. Thus, the triplet-based methods focused on relative distances, often have superior performance compared to pairwise algorithms. The margin-based Hinge loss function utilized in many triplet methods is defined as:

$$l((x_i, x_i^+, x_i^-)) = [\alpha - (d^- - d^+)]_+ = \begin{cases} 0, & if\ (d^- - d^+) \geq \alpha \\ \alpha - (d^- - d^+), & otherwise \end{cases} \quad (5)$$

Here, $d^+ = \|f_W(x_i) - f_W(x_i^+)\|$ indicates the Euclidean distance in the embedding space between anchor and positive and $d^- = \|f_W(x_i) - f_W(x_i^-)\|$ shows the distance between anchor and negative.

Most research in DML is focused on developing a new sampling strategy and loss functions. In the following, we review seminal work in each field.

## 2.1 Sampling Strategies

The simplest approach to generate positive and negative pairs is to randomly sample pairs based on class labels. For example, we can generate a similar pair by randomly selecting two datapoints from the same class and create a dissimilar pair by choosing two data items from different classes. This approach named easy sampling (Kaya and Bilge 2019) often results in poor performance. (Simo-Serra, Trulls et al. 2015) proposed hard negative mining that selects $x_i^-$ from opposite classes subject to $d^- < d^+$ condition. Facenet (Schroff, Kalenichenko et al. 2015) choose a *semi-hard* negative instance for each positive pair $(x_i, x_i^+)$. Here, the negative has more distance from the anchor compared to the positive example but still violates the margin constraint $(i.e., d^+ < d^-, (d^- - d^+) < \alpha)$. Figure 3 illustrates the differences between hard, semi-hard, and easy triplets.

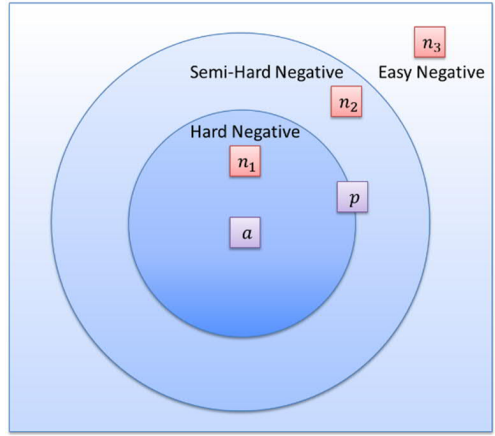

$a$: $anchor$, $p$: $positive$, $n_1$, $n_2$, $and$ $n_3$: $negative$

hard negative: $d^- < d^+$

semi-hard negative: $(d^- - d^+) < \alpha$

easy negative: $(d^- - d^+) \geq \alpha$

**Figure 3- Illustration of different triplet sampling strategies.**

Since hard negative samples are too close to the anchor, the gradient of loss has high variance and a low signal-to-noise ratio (Wu, Manmatha et al. 2017). To alleviate this issue, (Wu, Manmatha et al. 2017) proposed a distance weighting sampling approach that allows exploiting different types of triplets in a noisy environment.

N-pair (Sohn 2016) samples from all negative classes as shown in Figure 4. Hence, the loss function benefits from $N-1$ negative samples (one from each negative class) simultaneously that leads to a better convergence rate in comparison with triplet-based methods.

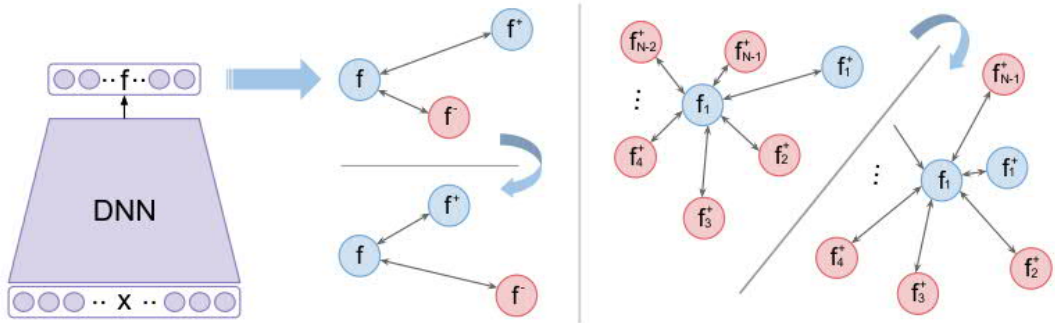

**Figure 4- Difference between triplet (left) and N-pair (right) distance learning. N-pair samples n-tuples that each contains an anchor ($f_1$), a positive example ($f_1^+$), and N-1 negative samples ($f_2^+,…,f_{N-1}^+$) (Sohn 2016).**

(Zheng, Chen et al. 2019) deals with the problem that many potential triplets in a dataset are easy and so do not affect the learned embedding. Hence, DML lacks enough informative samples for training. To overcomes the challenge, it generates a harder synthetic triplet from an easy sample ($\mathbf{z}, \mathbf{z}^+, \mathbf{z}^-$) by closing $\mathbf{z}^-$ to $\mathbf{z}$ while preserving its label information.

Generally, sampling algorithms lack different aspects such as capturing the global structure of data, low convergence rate, and extra time needed to mine informative samples. To address these challenges, proxy-based DML methods (Movshovitz-Attias, Toshev et al. 2017, Oh Song, Jegelka et al. 2017, Qian, Shang et al. 2019) proposed that can directly learn a semantic embedding from training data without utilizing a sampling strategy.

## 2.2 Loss Functions

A Loss function has a crucial role in the success of a DML algorithm. The *contrastive* and *triplet* are two widely used losses denoted in (1), and (2) equations, respectively. Many other losses are proposed to enhance the convergence rate and the discrimination power of the learned metric. (Ge 2018) adapts the margin in the triplet loss by constructing a hierarchical class-level tree. While this approach enhances the performance of DML, it suffers from the high computation time required to construct and update the tree.

The angular loss presented in (Wang, Zhou et al. 2017) constrains the *angle* at the negative instance of the triangle determined by a triplet. Compared to the contrastive and triplet losses, the loss benefits from robustness against feature variance and has a better convergence rate.

(Ni, Liu et al. 2017) aims to learn a better semantic embedding by sampling quadruplets instead of triplets. As shown in Figure 4, a quadruplet is formed by adding an extra positive example to a triplet, and the loss function uses two different margin values. The histogram loss (Ustinova and Lempitsky 2016) utilizes the histogram to compute a distribution-based similarity between positive and negative pairs. It then minimizes the probability that a positive pair has a lower similarity score than a negative.

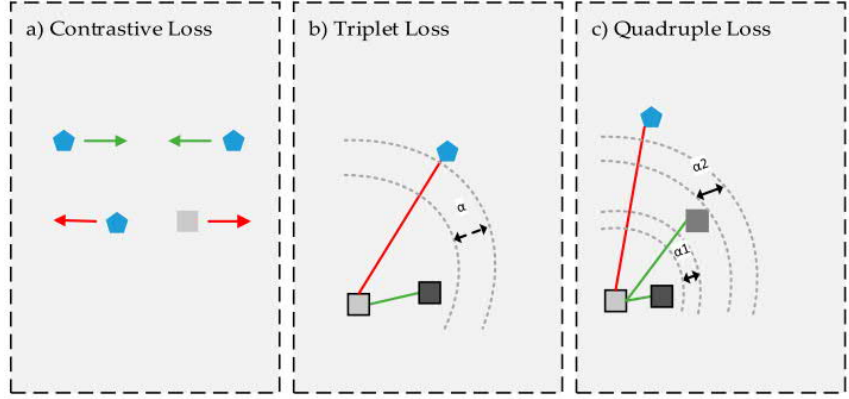

**Figure 5- The quadruplet loss vs contrastive and triplet loss**

(Wang, Han et al. 2019) proposes the GPW[1] mechanism that casts the sampling problem in DML into a pair weighting formulation. Also, it introduces the MS[2] loss under the GPW framework that explores pairs and weights them iteratively.

(Yao, Zhang et al. 2019) presents Part loss for the person re-identification task. It divides the identified body in a person image into five parts and then enforces the network to learn a representation for each different body part. By combining discriminative features obtained from different regions, the model achieves a better generalization to identify unseen persons.

For each similar pair, (Oh Song, Xiang et al. 2016) considers all negative samples within the input mini-batch. Here, the loss function attempts to keep away all negative samples from the positive. N-pair loss (Sohn 2016) defined in the input n-tuple as follows:

$$l_{n-pair}\left((x, x^+, x_1^-, x_2^-, \ldots, x_{N-1}^-)\right) = -\ln\left(1 + \sum_{j=1}^{N-1} \exp(S_j^- - S^+)\right) \qquad (6)$$

$$where, S^+ = f_w(x)^T f_w(x^+), \quad S_j^- = f_w(x)^T f_w(x_j^-)$$

Hence, the loss increases the similarity gap (i.e., $S^+ - S_j^-$) between a positive pair and all $N-1$ negatives simultaneously. Compared to the triplet loss, N-tuple losses achieve a higher convergence rate and better capture the global structure of semantic embedding by considering more training examples at a time.

Some recent work learns semantic embedding without the need to sampling tuples. For example, the Magnet loss (Rippel, Paluri et al. 2015) models distributions of classes in the embedding space and penalizes overlap between distributions. The clustering loss (Oh Song, Jegelka et al. 2017) directly optimizes the NMI[3]. As illustrated in Figure 6, it considers one proxy per class, and the loss brings examples to their proxies closer while penalizing different proxies to getting close to each other.

---

[1] General Pair Weighting

[2] Multi-Similarity

[3] Normalized Mutual Information

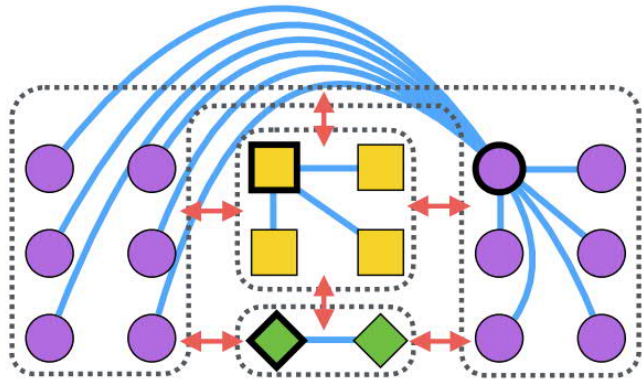

**Figure 6- (Oh Song, Jegelka et al. 2017) optimizes NMI clustering metric by considering one proxy per class.**

Proxy NCA (Movshovitz-Attias, Toshev et al. 2017) optimizes triplet loss where the triplets are formed by anchor data and proxies of positive and negative classes. Here, the proxies (one per class) are learned jointly with the semantic embedding. (Qian, Shang et al. 2019) shows that the classification loss is equivalent to a smoothed triplet loss where each class is represented by a single center. Then, it presents SoftTriple loss that extends classification loss by considering multiple proxies per class. It also provides an effective regularization term to determine the appropriate number of centers.

## 2.3 Generalizability of DML

Major applications of DML such as ZSL, FSL, person-reidentification, CBIR, and clustering are challenging due to the following characteristics:

1- A large number of classes
2- Few examples in some categories
3- Unseen classes during the test phase

Much research in DML focus on increasing the discrimination power of the learned embedding using available training classes and neglect the generalizability of the embedding on unseen categories.

MSML (Jiang, Huang et al. 2020) deals with the problem by utilizing both high-level semantic and low-level but abundant visual features. It learns multi-scale feature maps and a multi-scale relation generation network (MRGN).

To enhance the generalizability of DML, (Li, Yu et al. 2020) extends the triplet to a K-tuplet network where an anchor point deals with K negative examples at a time. That is similar to the test phase of an FSL that a query image needs to be compared with multiple different classes. The paper also provides a sampling mechanism to mine semi-hard informative K-tuples. However, these ideas coming from previous research such as (Sohn 2016) and mainly effective to capture the global structure of the embedding.

Adversarial learning is applied in DML to generate hard synthetic samples from the original training information. Unlike many approaches that ignore easy negatives, DAML[1] (Duan, Zheng et al. 2018) utilizes them to generate hard synthetic triplets. (Wang, Wang et al. 2020) presents a similar idea. Here, as illustrated in Figure 7, a hard negative generator is adversarially trained jointly with the DML to enhance the robustness of the learned embedding.

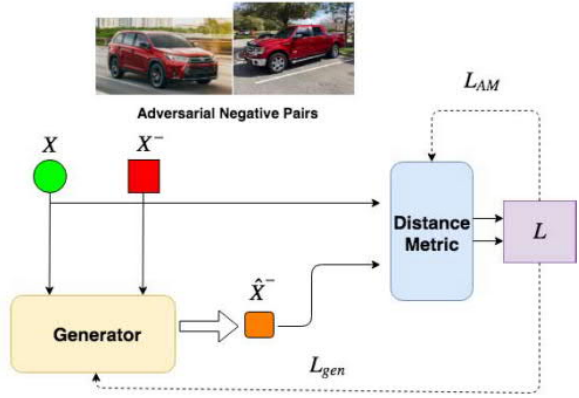

**Figure 7- Training distance metric and adversarial generator simultaneously to generate hard synthetic triplets (Wang, Wang et al. 2020)**

In (Xu, He et al. 2019) adversarial learning is exploited to enhance the performance of the semantic embedding in a multi-modal environment. To this end, it maps the text and corresponding image of a multi-modal data into a shared embedding where an adversarial classifier is employed to minimize the gap between these modalities (i.e., text and image).

(Chen and Deng 2019) proposes to use adversarial learning to enhance the performance of DML on unseen classes in the ZSL setting. It is obtained by introducing a regularization term named energy confusion that reduces overfitting on the seen classes during the metric learning process. The Same idea is also adopted in (Zhu, Zhong et al. 2020) to boost the discrimination power of an unconstrained palmprint recognition model. The energy confusion term is defined as:

$$l_{ec}(\boldsymbol{\theta}_f; X_I, X_J) = \mathbb{E}_{\tilde{X}_I, \tilde{X}_J}\left[\left\|\tilde{X}_I - \tilde{X}_J\right\|_2^2\right] = \sum_{i,j} p_{ij}\left\|x_i - x_j\right\|_2^2 \qquad (7)$$

where $\boldsymbol{\theta}_f$ denotes DML parameters. $X_I$ and $X_J$ are features from two different classes. $\tilde{X}_I$ and $\tilde{X}_J$ are random variables from probability distributions of $i^{th}$ and $j^{th}$ classes, respectively. $x_i$ and $x_j$ are corresponding observations from these classes. Also, $p_{ij}$ indicates joint probability distribution. Assuming independence between the classes and $\tilde{X}_I \sim Uniform(X_I), \tilde{X}_J \sim Uniform(X_J)$, $p_{ij}$ is:

$$p_{ij} = p_i p_j = \frac{1}{N_I}\frac{1}{N_J}$$

---

[1] Deep Adversarial Metric Learning

where $N_I$ and $N_J$ are the number of instances in the $i^{th}$ and $j^{th}$ classes. According to (7), minimizing $l_{ec}$ encourages overlap between two different classes that is in contradiction with traditional metric learning.

Our aim is also to promote the generalizability of DML. However, instead of using regularization terms, we utilize an adversarial classifier to increases the classification loss on seen classes during the semantic embedding learning. Besides, we learn global and discriminative features using feature maps of the intermediate layers and attention mechanisms.

## 3. Proposed Model

As illustrated in Figure 8, the proposed model, consists of four components: 1) Feature Extractor, 2) Tuplet Sampler, 3) Metric embedding layer, 4) Classifier.

First, the input image is passed through the deep network, and feature maps of the layers are generated. The feature maps of the selected intermediate layers are weighted using an attention mechanism and then combined to form the final feature vector. The extracted features of the input minibatch are fed to the tuplet sampler to generate training side information. The sampled tuples are forwarded to the metric embedding layer and the metric loss is evaluated. Besides, the feature vectors are given to the classifier to obtain the classification loss. The proposed hybrid loss function is evaluated, and the gradient is backpropagated to adjust the model parameters. The loss enforces the model to learn a discriminative semantic embedding that is not limited to observed categories and generalizes well on unseen classes. In the following, we discuss each component in more detail.

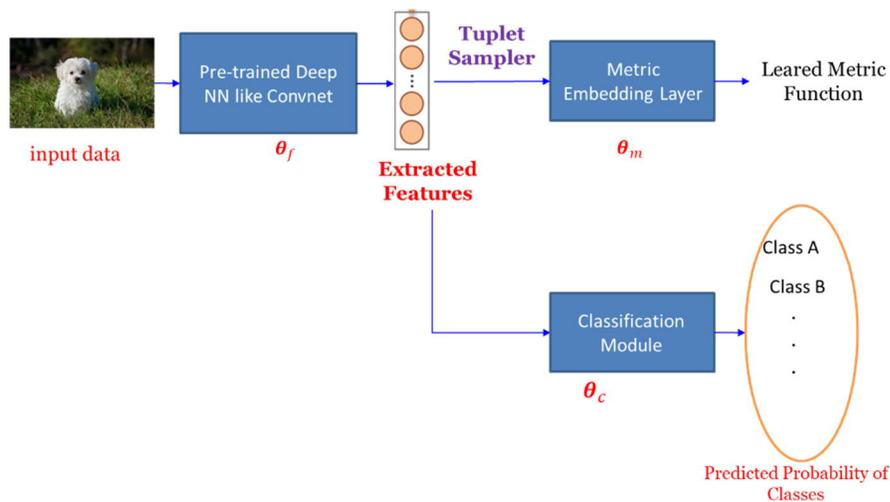

**Figure 8- The proposed model to enhance generalization of a DML Task.**

### 3.1 Feature Extractor

A deep neural network such as CNN, Encoder, LSTM, ... can be utilized to extract features from data. Since DML applications are focused on images, CNN models are very popular in this domain. A CNN learns hierarchical features from data. The initial layers extract more general and small patterns, and the last layers learn abstract discriminative concepts.

As mentioned, most DML algorithms only utilize the feature vector $\boldsymbol{u} \in \mathbb{R}^l$ of the last hidden layer from a pre-trained convnet such as InceptionV2, VGG19, or ResNet50. It increases the dependency of the learned metric on the seen classes. To resolve this issue, we divide the deep model into several sequential modules that increasingly learn discriminative and abstract patterns from the input. Figure 9 illustrates the proposed architecture.

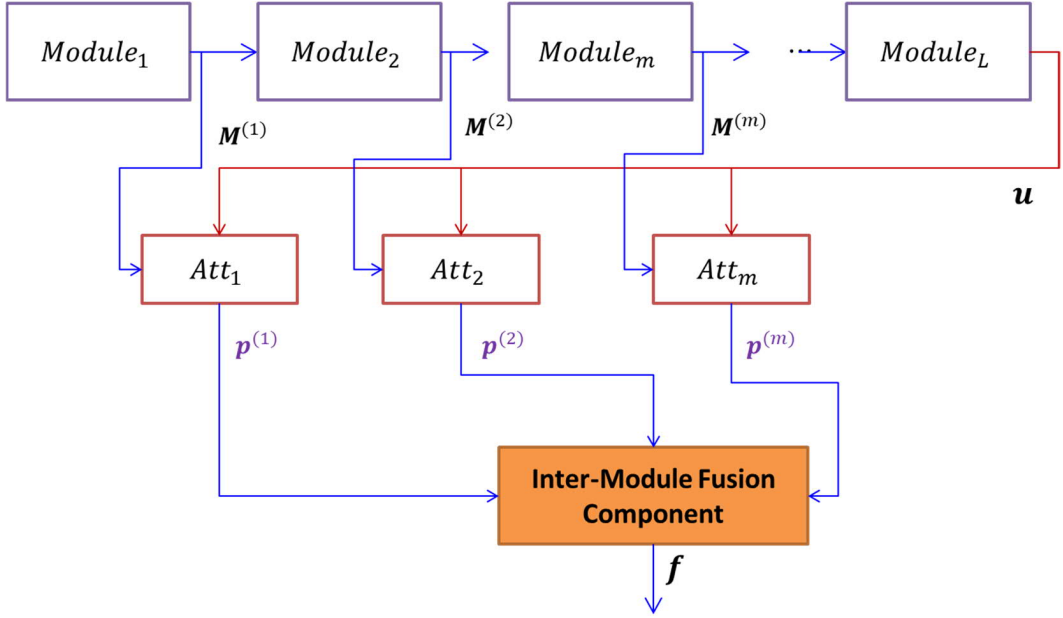

**Figure 9- The Architecture of feature extractor to learn global yet discriminative features from the input image.**

In Figure 9, $\boldsymbol{M}^{(i)} \in \mathbb{R}^{C_i \times H_i \times V_i}$ shows the output feature map from the i-th module. The task of the $Att_i$ is to attend features in $\boldsymbol{M}^{(i)}$ to the discriminative feature vector $\boldsymbol{u}$. The attention weight measures the discrimination power of the features. The final feature vector ($\boldsymbol{p}^{(i)} \in \mathbb{R}^l$) is formed by combining the weighted feature maps. Finally, the Inter-Module Fusion Component merges the attended features vectors. A straightforward mechanism to implement this module is to *concatenate* the attended vectors ($\boldsymbol{p}^{(i)}\ i = 1,2,\dots,m$). Some strategies to implement $Att_i$ are as follows.

I. Multiplicative Attention:

Let $\boldsymbol{q}_j \in \mathbb{R}^{c_i}$, $j = 1,2, \dots, (H_i \times V_i)$ be a channel in the feature map $\boldsymbol{M}^{(i)}$. In this approach, we measure the similarity of $\boldsymbol{q}_j$ to feature vector $\boldsymbol{u}$ as (Shen, Zhou et al. 2018):

$$S(\boldsymbol{q}_j, \boldsymbol{u}) = \langle \boldsymbol{W}_i^{(1)} \boldsymbol{q}_j, \boldsymbol{W}_i^{(2)} \boldsymbol{u} \rangle \tag{8}$$

In the above equation, the weight matrices $\boldsymbol{W}_i^{(1)}$ and $\boldsymbol{W}_i^{(2)}$ are learned in an end-to-end training paradigm to increase the similarity between $\boldsymbol{q}_j$ and $\boldsymbol{u}$. Since we aim to promote the similarity of $q_j$ to u, the weight matrix $W_i^{(2)}$ can be dropped to achieve a simpler (required fewer parameters) as follows:

$$S(q_j, u) = \langle W_i^{(1)} q_j + u \rangle \tag{9}$$

where $W_i^{(1)} \in \mathbb{R}^{l \times c_i}$.

II. Additive Attention

This mechanism evaluates the similarity of $q_j$ to $u$ (Shen, Zhou et al. 2018) as:

$$S(q_j, u) = w_i^T \sigma \left( W_i^{(1)} q_j + W_i^{(2)} u \right) \tag{10}$$

where $\sigma$ is an activation function and $w_i$ is a weight vector. In practice, the additive form performs better than the multiplicative mechanism. However, it demands more computations and memory. Similarly, we can change the above relation to a simpler form as:

$$S(q_j, u) = w_i^T \sigma \left( W_i^{(1)} q_j + u \right) \tag{11}$$

III. Multi-dimensional Attention (Shen, Zhou et al. 2018):

This mechanism replaces the weight vector $w_i$ with a matrix $W_i \in \mathbb{R}^{l \times c_i}$.

$$S(q_j, u) = W_i^T \sigma \left( W_i^{(1)} q_j + u \right) \tag{12}$$

Thus, the output is a similarity vector $a_j \in \mathbb{R}^{c_i}$ where each comonent in $a_j$ measures the importance or weight of corresponding feature in $q_j$. After computing the similarity function, the softmax is applied to normalize attention weights. More precisely, let $a^{(i)} = [S(q_j, u)]_{j=1}^{H_i \times V_i}$ be the attention vector obtained from module $i$ using multiplicative or additive mechanisms. We normalize the $a^{(i)}$ using the softmax function as follows:

$$\tilde{a}_j^{(i)} = p(z_j | q_j, u) = [\text{softmax}(a^{(i)})]_j = \frac{\exp(S(q_j, u))}{\sum_{k=1}^{H_i \times V_i} \exp(S(q_j, u))} \tag{13}$$

Here, $z_j$ is an indicator that shows which channel in the i-th module is attended more to the reference vector $u$. In other words, the $\tilde{a}_j^{(i)} = p(z_j | q_j, u)$ denotes the discrimination score or attention weight of $q_j$. The output of the attention is the weighted mean of channels:

$$p^{(i)} = \sum_{j=1}^{H_i \times V_i} \tilde{a}_j^{(i)} q_j, \quad i = 1, 2, \ldots, m \tag{14}$$

In the case of multi-dimensional attention, we perform the same procedure to each feature independently to obtain the importance of the feature. In particular, let $A_{jk}$ be the k-th component of the $S(q_j, u)$ vector. $A_{jk}\ k = 1,2,\ldots,c_i$ shows the unnormalized weight of k-th feature in $q_j$. We normalize the weights using the Softmax as:

$$\widetilde{A_{jk}}^{(i)} = p(z_{jk}|q_j, u) = \frac{\exp(A_{jk})}{\sum_{l=1}^{c_i} \exp(A_{jl})} \quad (15)$$

Then, the output of the k-th component of $p^{(i)}$ is obtained as follows:

$$p_k^{(i)} = \sum_{j=1}^{H_i \times V_i} \widetilde{A_{jk}}^{(i)} q_{jk}, \quad i = 1,2,\ldots,m, \quad k = 1,2,\ldots,c_i \quad (16)$$

In the following, we show the Feature Extractor as a function $f(\theta_f; .)$ where $\theta_f$ denotes the network parameters.

### 3.2 Tuplet Sampler

As mentioned, many DML algorithms use pairs or triplet constraints. A triplet sample considers both negative and positive constraints at the same time. Also, the triplet loss focuses on relative distances while contrastive loss concentrates on absolute distances. Therefore, triplet-based methods often outperform pair-based algorithms. Some work extends the triplet sampling to quadruplets (Ni, Liu et al. 2017) and N-pairs (Sohn 2016). Recently, proxy-based methods are introduced that aim to learn the semantic embedding without a sampling procedure (Movshovitz-Attias, Toshev et al. 2017, Qian, Shang et al. 2019).

As seen in Figure 8, the proposed approach can be easily applied to any DML method. In our work, the sampling procedure is selected based on the DML algorithm. The input of this algorithm is a mini-batch $B$ of extracted features along with their labels.

$$B = \{((f_1, y_1), (f_2, y_2), \ldots, (f_n, y_n))\}, \text{ where } f_i = f(\theta_f; x_i)$$

The output is a set of sampled tuples denoted by:

$$T = \{T_1, T_2, \ldots, T_N\}$$

Note that proxy-based DML methods do not need a sampling procedure. In this case, we directly pass the mini-batch $B$ to the metric embedding layer.

### 3.3 Metric Embedding Layer

This layer is built on top of the feature extractor. The input of the layer is a mini-batch of tuples generated by the sampler. As our approach can be applied to almost any DML method, we can generally denote this layer by the function $g(\theta_m; .)$ that transforms the extracted features to the semantic embedding space.

Let $l_m$ be the metric loss function, we can optimize the parameters $\{\theta_f, \theta_m\}$ by solving the following optimization problem:

$$\widehat{\boldsymbol{\theta}}_f, \widehat{\boldsymbol{\theta}}_m = \arg \min_{\boldsymbol{\theta}_f, \boldsymbol{\theta}_m} \frac{1}{N} \sum_{i=1}^{N} l_m(\boldsymbol{T}_i; \boldsymbol{\theta}_f, \boldsymbol{\theta}_m) \tag{17}$$

Minimizing the above loss function helps to learn an appropriate semantic embedding space for the classes included in the training dataset. However, it may not be suitable for the new classes (or even observed categories with few samples). Thus, we should force the feature extractor to mine a more general representation and discard the features specific to the observed classes. To this end, we propose to combine the DML process with the classification module in an adversarial manner.

### 3.4 Classification Module

This module can be implemented by a multi-layer neural network with the Softmax activation function at the last layer. Generally, we can denote this module by $\boldsymbol{h}(\boldsymbol{\theta}_c; \boldsymbol{f}_i)$ where $\boldsymbol{\theta}_c$ is the parameters and $\boldsymbol{f}_i$ is the extracted feature vector from the input image (i.e., $\boldsymbol{f}_i = \boldsymbol{f}(\boldsymbol{\theta}_f, \boldsymbol{x}_i)$). Cross-Entropy is the most popular classification loss used in many models defined as:

$$l_{cross-entropy}(\boldsymbol{x}_i, y_i; \boldsymbol{\theta}_f, \boldsymbol{\theta}_c) = -\log p_{y_i} = -\log \sum_{k=1}^{C} \mathbf{1}_{(y_i=k)} [\boldsymbol{h}(\boldsymbol{\theta}_c; \boldsymbol{f}_i)]_k \tag{18}$$

where $p_{y_i}$ is the probability of the correct label obtained by applying the Softmax on the last layer of the module and $[\boldsymbol{h}(\boldsymbol{\theta}_c; \boldsymbol{f}_i)]_k$ denotes the k-th component of the classification output. Generally, we indicate the classification loss by $l_c(\boldsymbol{x}_i; \boldsymbol{\theta}_f, \boldsymbol{\theta}_c)$.

### 3.5 Model Integration

In the training process, we aim to learn a semantic embedding space by minimizing the metric loss (17) on the $\{\boldsymbol{\theta}_f, \boldsymbol{\theta}_m\}$ parameters. Meanwhile, we should enforce the feature extractor to discard class-specific features by utilizing the classification module. To this end, we propose two alternative adversarial approaches:

I. Label Smoothing Approach

This approach replaces a one-hot label $\boldsymbol{y}^{(hot)}$ with a probability vector formed by the combination of $\boldsymbol{y}^{(hot)}$ with a uniform distribution. It can be formulated as:

$$\boldsymbol{y}^{(ls)} = (1 - \alpha)\boldsymbol{y}^{(hot)} + \frac{\alpha}{C} \tag{19}$$

Here, $\alpha$ is the smoothing factor and $C$ is the number of available training classes. Label smoothing is an effective means to prevent both overfitting and overconfidence problems. Also, it allows the feature extractor to learn a more general representation from the inputs by relaxing the Cross-Entropy loss. The classification loss in this approach is as follows:

$$l_c(\boldsymbol{x}_i; \boldsymbol{\theta}_f, \boldsymbol{\theta}_c) = -\sum_{k=1}^{C} y_{ik}^{(ls)} \log[\boldsymbol{h}(\boldsymbol{\theta}_c; \boldsymbol{f}_i)]_k \tag{20}$$

where $y_{ik}^{(ls)}$ denotes the k-th component of the label $\boldsymbol{y}_i^{(ls)}$. Finally, we formulate the proposed loss function in this approach as:

$$l_{s\_adv}(B, T; \boldsymbol{\theta}_f, \boldsymbol{\theta}_m, \boldsymbol{\theta}_c) = l_m(T, \boldsymbol{\theta}_f, \boldsymbol{\theta}_m) + \lambda\, l_c(B, \boldsymbol{\theta}_f, \boldsymbol{\theta}_c) \tag{21}$$

Here, the hyper-parameter $\lambda$ balances the trade-off between the metric and classification losses.

II. Adaptive Adversarial Approach

In this approach, we utilize the classification module in an adaptive adversarial setting. We observed that in the initial stages of training, minimizing the classification loss on seen classes helps to learn a better discriminative representation. However, in the later stages, it is necessary to *increase* the classification loss (*jointly with minimizing the metric loss*) to enforce the feature extractor to learn more general discriminative features that are not limited to available classes in the training set. Therefore, we propose the final loss function as:

$$l_{adapt-adv}(B, T; \boldsymbol{\theta}_f, \boldsymbol{\theta}_m, \boldsymbol{\theta}_c) = l_m(T, \boldsymbol{\theta}_f, \boldsymbol{\theta}_m) - \lambda\, l_c(B, \boldsymbol{\theta}_f, \boldsymbol{\theta}_c) \tag{22}$$

When $\lambda < 0$, the proposed loss is optimized by minimizing both the metric and classification loss on seen classes. After a certain classification loss is met, we change the sign of $\lambda$ and force the optimizer to increase the classification loss.

To optimize $l_{adapt-adv}$, we first freeze the classifier parameters (i.e., $\boldsymbol{\theta}_c$) and minimize the loss on the feature extractor and metric embedding layer parameters as follows:

$$\widehat{\boldsymbol{\theta}}_f, \widehat{\boldsymbol{\theta}}_m = \arg\min_{\boldsymbol{\theta}_f, \boldsymbol{\theta}_m} R_{adv} = \frac{1}{N}\sum_{i=1}^{N} l_m(T_i; \boldsymbol{\theta}_f, \boldsymbol{\theta}_m) - \frac{\lambda}{n}\sum_{i=1}^{n} l_c(x_i; \boldsymbol{\theta}_f, \boldsymbol{\theta}_c) \tag{23}$$

Subsequently, the classifier tries to minimize the classification loss by maximizing the loss as follows:

$$\widehat{\boldsymbol{\theta}}_c = \arg\max_{\boldsymbol{\theta}_c} R_{adv} \equiv \arg\min_{\boldsymbol{\theta}_c} \frac{\lambda}{n}\sum_{i=1}^{n} l_c(x_i; \boldsymbol{\theta}_f, \boldsymbol{\theta}_c) \tag{24}$$

It motivates a minimax game between the classifier and the feature extractor. On one hand, the feature extractor confuses the classifier by maximizing $l_c$, and on the other hand, the classifier tries to find class-discriminative information in the feature representations to identify the correct label. Algorithm 1 summarizes the steps of the proposed adversarial approach.

In both approaches, utilizing dropout layers in the classification module is appropriate. The dropout mechanism simulates a large number of different network architectures by randomly dropping some nodes at the training phase. Thus, relaxing or increasing the classification loss on the simulated classifiers leads to obtain a more generalizable representation. Besides, Dropout is an effective yet cheap regularization technique that avoids overfitting and decreases the generalization error of deep neural networks.

| Algorithm1. The proposed adversarial approach |
|---|
| Inputs: $\{(x_i, y_i), i = 1, 2, \ldots, L\}$, $\lambda$ : Controls trade-off between metric learning and classifier objectives |
| Output: Learned Embedding $h$, Feature Extractor $f$ |

    1. Initialize the feature extractor with a pre-trained Convnet

    2. for iter $= 1, 2, \ldots MAX\_Iter$

        2.1. B← Get a mini-batch of data and forward it through the network to generate features vectors.

        2.2. Obtain the training tuples $T$ from the input batch $B$ using the sampling module[1].

        2.3. Optimize deep feature extractor and metric embedding parameters:

$$\widehat{\boldsymbol{\theta}}_f, \widehat{\boldsymbol{\theta}}_m = \arg\min_{\boldsymbol{\theta}_f, \boldsymbol{\theta}_m} R_{adv} = \frac{1}{N}\sum_{i=1}^{N} l_m(\boldsymbol{T}_i; \boldsymbol{\theta}_f, \boldsymbol{\theta}_m) - \frac{\lambda}{n}\sum_{i=1}^{n} l_c(\boldsymbol{x}_i; \boldsymbol{\theta}_f, \boldsymbol{\theta}_c)$$

        using sophisticated neural network optimization algorithms like Adam.

        2.4. Optimize the classifier module:

$$\widehat{\boldsymbol{\theta}}_c = \arg\max_{\boldsymbol{\theta}_c} R_{adv} \equiv \arg\min_{\boldsymbol{\theta}_c} \frac{\lambda}{n}\sum_{i=1}^{n} l_c(\boldsymbol{x}_i; \boldsymbol{\theta}_f, \boldsymbol{\theta}_c)$$

        using sophisticated neural network optimization algorithms like Adam

    end;

## 4. Implementation Details

To implement the proposed approaches, we utilize the pre-trained *Inception v2* neural network (Szegedy, Vanhoucke et al. 2016) as the feature extractor[2].

    The network consists of 5 inception modules. Figure 10-(a) illustrates the architecture of each module. Here, we utilize the feature maps of intermediate layers *inception_4d_output, inception_4e_output, and inception_5a_output.* The size of feature maps in these layers are $608 \times 14 \times 14$, $1056 \times 7 \times 7$, and $1024 \times 7 \times 7$, respectively. The output of the last hidden layer is $\boldsymbol{u} \in \mathbb{R}^{1024}$. We use the *additive* attention mechanism denoted in equation (11) to promote the discrimination power of the intermediate features. The output feature vectors of attention modules are then *concatenated* in the *Inter-Module Fusion Component*. The Simplified architecture of the implemented feature extractor is shown in Figure 10-(b).

---

[1] For proxy-based DML, we directly pass the mini-batch $B$ to the metric embedding layer.

[2] downloaded from: https://github.com/dichotomies/proxy-nca

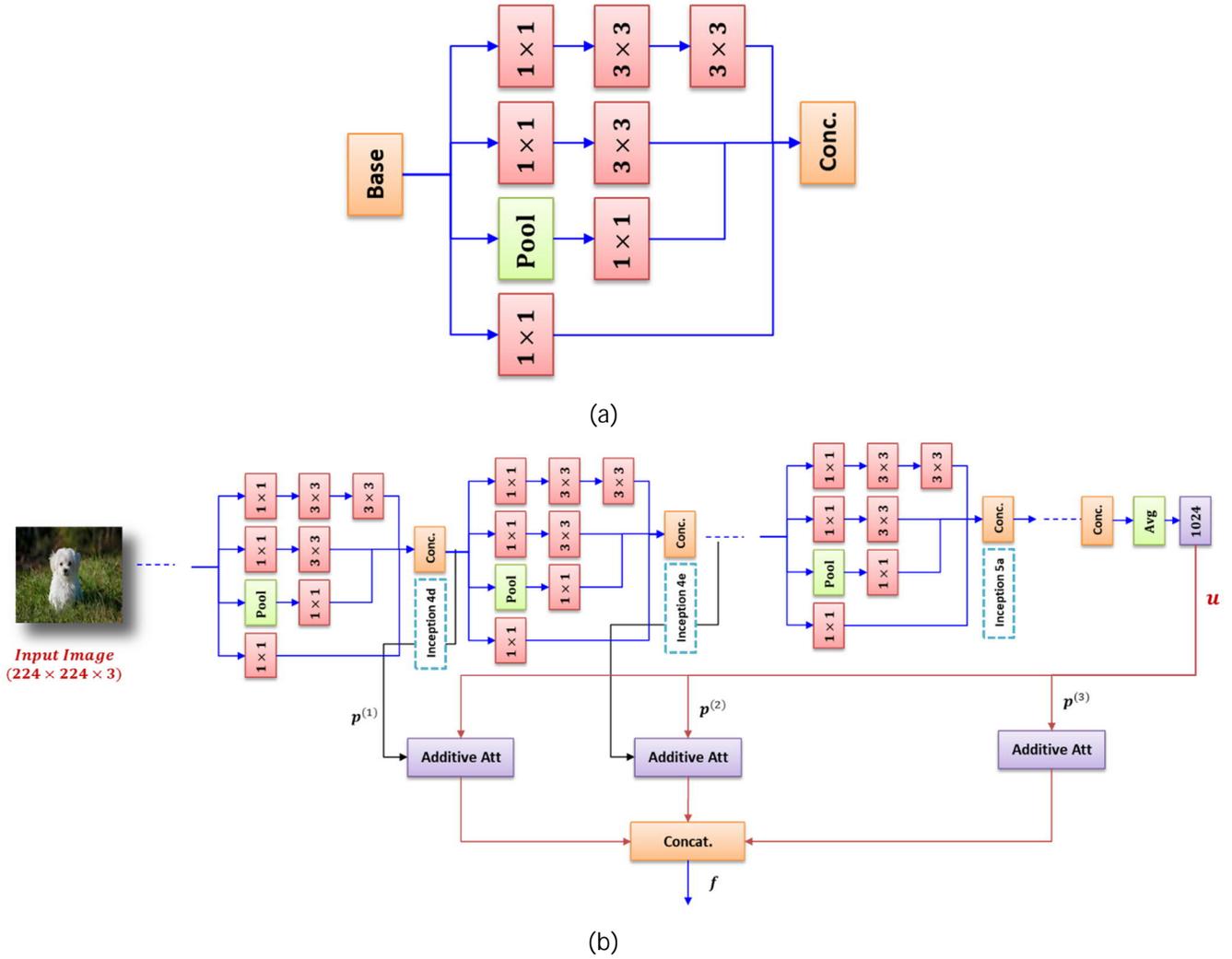

**Figure 10- (a)-Inception module architecture in the *Inception V2*, (b) The Simplified architecture of the implemented feature extractor**

To implement the adaptive adversarial approach, we used a *GRL*[1] layer between the classification module and the feature extractor. The GRL acts as an identity operation in the forward phase. However, during the backward stage, it multiplies the gradient with $-\lambda$ and then passes the results to the previous layer. We adjust the $\lambda$ based on the classification loss ($l_c$) in each epoch as follows:

$$\lambda = -\tanh(l_c - l_{thresh})\lambda_0 \tag{25}$$

where $l_{thresh}$ indicates a classification loss threshold. Hence, in the initial training epochs where $l_c$ is high, the $\lambda$ is adjusted with a negative value. As the $l_c$ is decreased substantially (i.e., $l_c < l_{thresh}$), the adversarial mechanism is activated and tries to prevent overfitting on the seen classes. The value of $\lambda_0$ is chosen from the range [.1, 1] as described in the experimental section.

---

[1] Gradient Reversal Layer

The classification module is implemented by a standard feed-forward neural network with one hidden layer followed by a dropout operation. The architecture of the module is shown in the following figure.

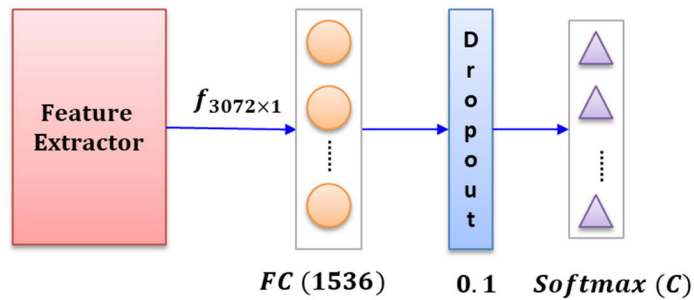

**Figure 11- Network Architecture of the Classification Module**

The work is implemented using *Pytorch* deep learning library and the source code is publicly available at: https://github.com/d-zabihzadeh/Adaptive-Adv-DML.

## 5. Experimental Results

In this section, we evaluate our work on some challenging datasets in the machine vision domain. To this end, we select Triplet hinge loss[1], Angular Loss[2] (Wang, Zhou et al. 2017), and Proxy-NCA[3] (Movshovitz-Attias, Toshev et al. 2017) as baseline DML algorithms and examine the significance of applying our framework to these methods.

According to the selected adversarial approach, we name the proposed methods as Soft-Adv-DML[4] and Adapt-Adv-DML[5] . Soft-Adv utilizes the label smoothing approach to relax the classification loss whereas Adapt-Adv employs the classification module adversarially to prevents overfitting on the observed classes. Also, we compare our work with the recent Energy confusion approach (Chen and Deng 2019) on the evaluated datasets. Subsequently, Hyper-parameter analysis and the ablation study of components in our methods are provided. We also visualize the attention weights of the feature maps that indicate which parts of the input images are attended by extracted feature maps.

### 5.1 Data Description

The CUB-200-2011, Cars, and Flower102 are widely used datasets that are selected in our work. The statistics of these datasets are summarized in Table 1.

---

[1] Source code: https://github.com/KevinMusgrave/pytorch-metric-learning

[2] Source code: https://github.com/tomp11/metric_learning

[3] Source code: https://github.com/dichotomies/proxy-nca

[4] Soft Adversarial Deep Metric Learning

[5] Adaptive Adversarial Deep Metric Learning

**Table 1- Statistics of the image datasets used in our experiments.**

| Data Set | #classes | #samples | Evaluation Protocol |
|---|---|---|---|
| CUB-200-2011 (Wah, Branson et al. 2011) | 200 | 11,788 | The initial 100 classes (including 5864 images) for training and the rest (100) for testing (5,924 images) |
| CARS-196 (Krause, Stark et al. 2013) | 196 | 16,185 | The first 98 classes (including 8,054 images) for training and the remaining ones (98) for testing (8,131 images) |
| Oxford Flowers-102 (Nilsback and Zisserman 2008) | 102 | 6,552 | The first 51 classes (including 2,807 images) for training and the remaining ones (51) for evaluation (3,745 images) |
| Online Products (Oh Song, Xiang et al. 2016) | 22,634 | 120,053 | The initial 11,318 classes (including 59,551 images) for training and the remaining categories (11,316) for testing (60,502 images) |

## 5.2 Evaluation Metrics

To evaluate the proposed methods, we adopt standard information retrieval metrics: Recall@$k$ and *NMI*[1]. Recall@$k$ measures the proportion of relevant images in the top-$k$ retrieved results.

*NMI* indicates the quality of clustering. Let $C = \{c_1, c_2, \dots, c_n\}$ be the clustering assignment set provided by a clustering method. Given the true labels $Y = \{y_1, y_2, \dots, y_n\}$, NMI is computed as:

$$NMI = 2 \times \frac{I(Y;C)}{H(Y) + H(C)} \tag{26}$$

where $I(;)$ is the mutual information, and $H(.)$ indicates the entropy.

Also, we evaluate the accuracy of kNN ($k=5$) in the ZSL setting. Here, a test image is classified correctly provided *3 or more* relevant (with the same label) images be among the *5-top* results.

## 5.3 Experimental Setup

For a fair comparison, we adopt the pre-trained *Inception-V2* for all evaluated methods. We found that the learning rate=$10^{-4}$ in the network is appropriate for fine-tuning. Thus, we set $lr = 10^{-4}$ for all evaluated methods. Also, similar to (Chen and Deng 2019), the embedding layer learning rate is set to $10^{-3}$ (10 times faster than other layers).

Like (Movshovitz-Attias, Toshev et al. 2017), the learned embedding size is set to 64, and we adopt the same transformations on the input images. The images are reshaped to $256 \times 256$, and then randomly cropped to $224 \times 224$. The batch-size is also set to 64 for all methods.

As our work concentrates on the improvement of DML on unseen classes, we almost keep the default values of hyper-parameters in DML methods for all datasets and perform small adjustments. For triplet-hinge DML, we adopt *'semi-hard'* sampling strategy. We also find out setting $margin = 0.01$ is appropriate on all the evaluated datasets. In Angular loss, we keep the default value of $\alpha = 45°$ for all methods. Also, the triplets are generated based on N-pair sampling (same as the provided source code). The learning rate of proxy centers in Proxy-NCA is selected from the range $\{0.01, 0.15, 0.02\}$.

---

[1] Normalized Mutual Information

Finally, in our methods, we use *smoothing factor* = 0.15, *dropout-rate* = 0.1, and $l_{thresh} = 1.5$, in all experiments. Also, the value of $\lambda_0$ and $\lambda$ is selected from $\{0.1, 0.5, 1\}$. The following table summarizes the hyper-parameters of evaluated DML along with their adjustments.

Table 2- Specifications of Hyperparameters of DML methods and their adjustments

| Hyper-parameter | Description | Value |
| --- | --- | --- |
| $model - lr$ | Learning rate of pre-trained Inception-V2 network. | $10^{-4}$ |
| $embedding - lr$ | Learning rate of the embedding layer. | $10^{-3}$ |
| $margin$ | The margin of triplet sampling and hinge loss | 0.01 |
| $\alpha$ | The degree in Angular loss | $45°$ |
| $proxy - lr$ | The learning rate of the Proxy-NCA | Adjusted from $\{0.01, 0.15, 0.02\}$ |
| smoothing factor | The label smoothing factor of the Soft-Adv DML | 0.15 |
| dropout rate | The dropout rate in the classification module | 0.1 |
| $\lambda_0$ and $\lambda$ | Balances the trade-off between the metric and classification losses in Adv-DML and Soft-Adv-DML respectively | Adjusted from $\{0.1, 0.5, 1\}$ |
| $l_{thresh}$ | The classification loss threshold | 1.5 |

### 5.4 Results and Analysis

In the first experiment, we evaluate the proposed methods in the CUB-200-2011, Cars-196, and Oxford Flowers-102 datasets in the ZSL setting as described in Table 1. The results are reported in Table 3, Table 4, and Table 5. respectively. Also, we visualize the *NMI* and Recall@$k$ scores of the evaluated methods versus epochs on the test classes in CUB-200-2011 and Cars-196 datasets in Figure 12 and Figure 13, respectively.

**Table 3- Information Retrieval Results in a ZSL Setting on the CUB-200-2011 dataset.**

| Method | Extension | NMI | Recall@1 | Recall@2 | Recall@4 | Recall@8 | kNN-Acc |
|---|---|---|---|---|---|---|---|
| Triplet Loss: | Base | 58.17 | 45.39 | 58.86 | 70.41 | 80.44 | 49.14 |
| | Energy | 55.87 | 39.94 | 52.65 | 66.12 | 78.53 | 42.35 |
| | Soft-Adv | 61.83 | 51.5 | 64.47 | 75.83 | 84.69 | 53.98 |
| | Adapt-Adv | 62.07 | 53.66 | 65.83 | 76.65 | 84.99 | 56.14 |
| Angular Loss: | Base | 59.26 | 49.59 | 62.15 | 73.36 | 83.05 | 51.98 |
| | Energy | 59.91 | 49.95 | 62.19 | 72.75 | 82.63 | 51.92 |
| | Soft-Adv | 61.65 | 52.18 | 64.96 | 75.79 | 84.47 | 54.90 |
| | Adapt-Adv | 61.78 | 52.68 | 65.11 | 75.86 | 84.5 | 55.60 |
| Proxy-NCA: | Base | 63.62 | 54.02 | 66.83 | 78.07 | 85.7 | 56.74 |
| | Energy | 63.62 | 53.83 | 66.49 | 77.67 | 86.38 | 56.48 |
| | Soft-Adv | 66.26 | 58.1 | 69.95 | 79.46 | 87.22 | 60.08 |
| | Adapt-Adv | 65.86 | 57.38 | 68.91 | 78.58 | 87.07 | 60.20 |

**Table 4- Information Retrieval Results in a ZSL Setting on the CARS-196 dataset.**

| Method | Extension | NMI | Recall@1 | Recall@2 | Recall@4 | Recall@8 | kNN-Acc |
|---|---|---|---|---|---|---|---|
| Triplet Loss: | Base | 52.50 | 44.13 | 57.51 | 70.1 | 79.79 | 46.28 |
| | Energy | 48.93 | 34.93 | 48.49 | 61.8 | 74.16 | 37.57 |
| | Soft-Adv | 57.61 | 61.74 | 73.36 | 82.52 | 89.51 | 63.82 |
| | Adapt-Adv | 57.71 | 62.51 | 74.23 | 83.22 | 89.45 | 65.18 |
| Angular Loss: | Base | 55.49 | 59.76 | 71.16 | 79.79 | 86.87 | 61.33 |
| | Energy | 55.52 | 60.34 | 71.37 | 80.09 | 87.27 | 62.29 |
| | Soft-Adv | 59.23 | 67.43 | 77.4 | 84.68 | 90.35 | 68.95 |
| | Adapt-Adv | 59.49 | 66.99 | 77.58 | 85.39 | 90.63 | 68.33 |
| Proxy-NCA: | Base | 61.20 | 65.67 | 76.67 | 84.68 | 90.65 | 67.78 |
| | Energy | 60.16 | 64.11 | 75.18 | 83.91 | 89.69 | 66.40 |
| | Soft-Adv | 63.90 | 70.84 | 79.68 | 87.04 | 92.09 | 72.52 |
| | Adapt-Adv | 63.53 | 69.94 | 79.66 | 86.37 | 91.42 | 71.76 |

**Table 5- Information Retrieval Results in a ZSL Setting on the on the Oxford 102 Flowers dataset.**

| Method | Extension | NMI | Recall@1 | Recall@2 | Recall@4 | Recall@8 | kNN-Acc |
|---|---|---|---|---|---|---|---|
| Triplet Loss: | Base | 70.16 | 81.55 | 88.89 | 93.59 | 96.58 | 82.22 |
| | Energy | 68.42 | 78.24 | 86.28 | 92.02 | 96.05 | 80.24 |
| | Soft-Adv | 76.13 | 89.61 | 94.10 | 97.04 | 98.56 | 90.63 |
| | Adapt-Adv | 76.59 | 89.32 | 94.26 | 96.48 | 97.84 | 90.39 |
| Angular Loss: | Base | 72.39 | 84.89 | 90.89 | 94.63 | 97.12 | 86.17 |
| | Energy | 71.39 | 85.21 | 90.81 | 94.31 | 96.8 | 85.93 |
| | Soft-Adv | 75.65 | 89.91 | 94.31 | 96.56 | 98.05 | 91.00 |
| | Adapt-Adv | 76.05 | 89.64 | 93.94 | 96.56 | 98.16 | 90.31 |
| Proxy-NCA: | Base | 73.79 | 86.3 | 91.24 | 95.33 | 97.38 | 88.38 |
| | Energy | 75.42 | 87.42 | 92.26 | 95.59 | 97.68 | 88.44 |
| | Soft-Adv | 79.77 | 91.05 | 94.98 | 97.25 | 98.42 | 92.15 |
| | Adapt-Adv | 80.46 | 91.67 | 95.27 | 97.62 | 98.69 | 92.42 |

As the results indicate, the proposed methods provide a considerable improvement over the baseline DML methods. It confirms that our approaches (general and discriminative feature learning + adversarial classification module) are indeed beneficial for the generalization of DML methods in a ZSL setting. They learn a more generalizable and discriminative representation from the input image and effectively prevent overfitting on the observed classes. Also, the obtained results of two proposed adversarial approaches are very competitive and none of them is significantly superior to the other.

Meanwhile, we did not observe significant advantage from the Energy confusion approach over the baseline methods. The main drawback of the approach is that the confusion coefficient $\lambda$ is fixed during the training phase. Hence, it prevents the baseline DML to learn the discriminative features efficiently during the initial stages of training. Besides, the energy confusion term does not consider the structure of the embedding and randomly selects instances from opposite classes. In contrast, our methods capture the global structure of the embedding by employing a classification module.

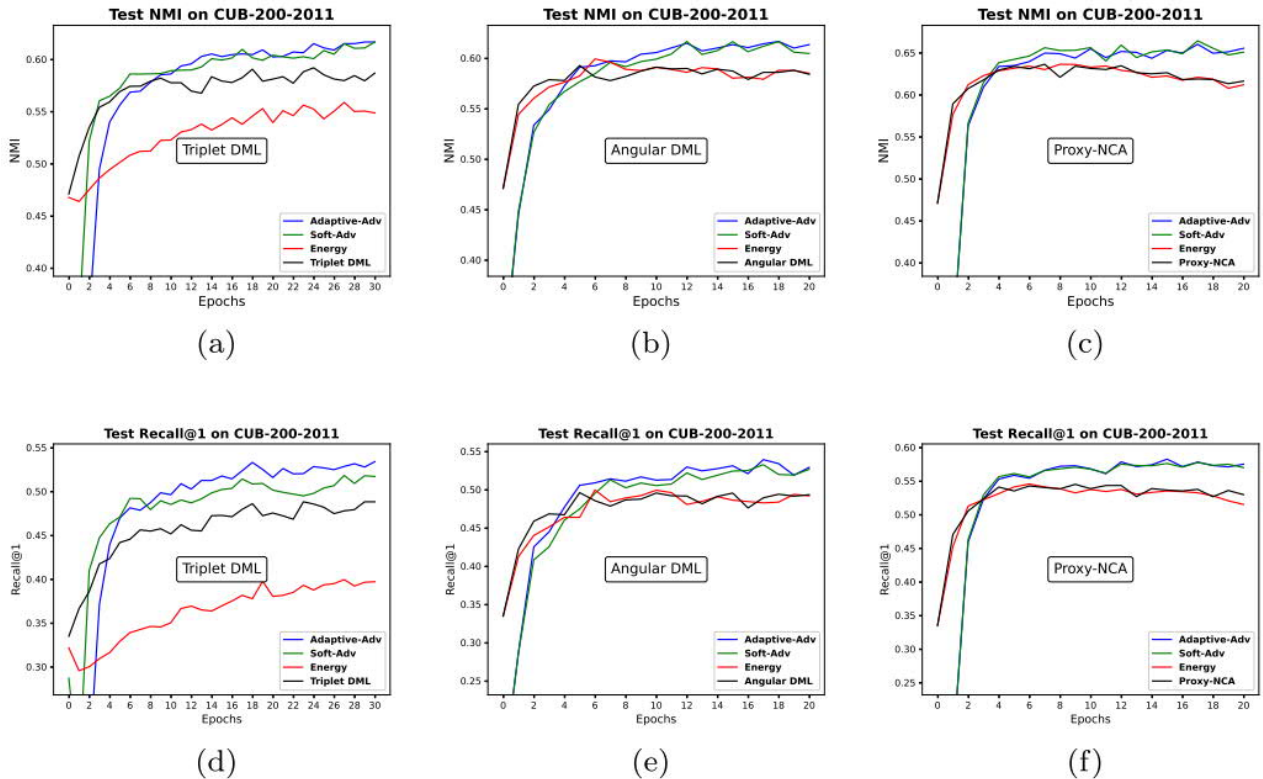

**Figure 12- NMI and Recall@1 of the evaluated methods on the CUB-200-2011 dataset.**

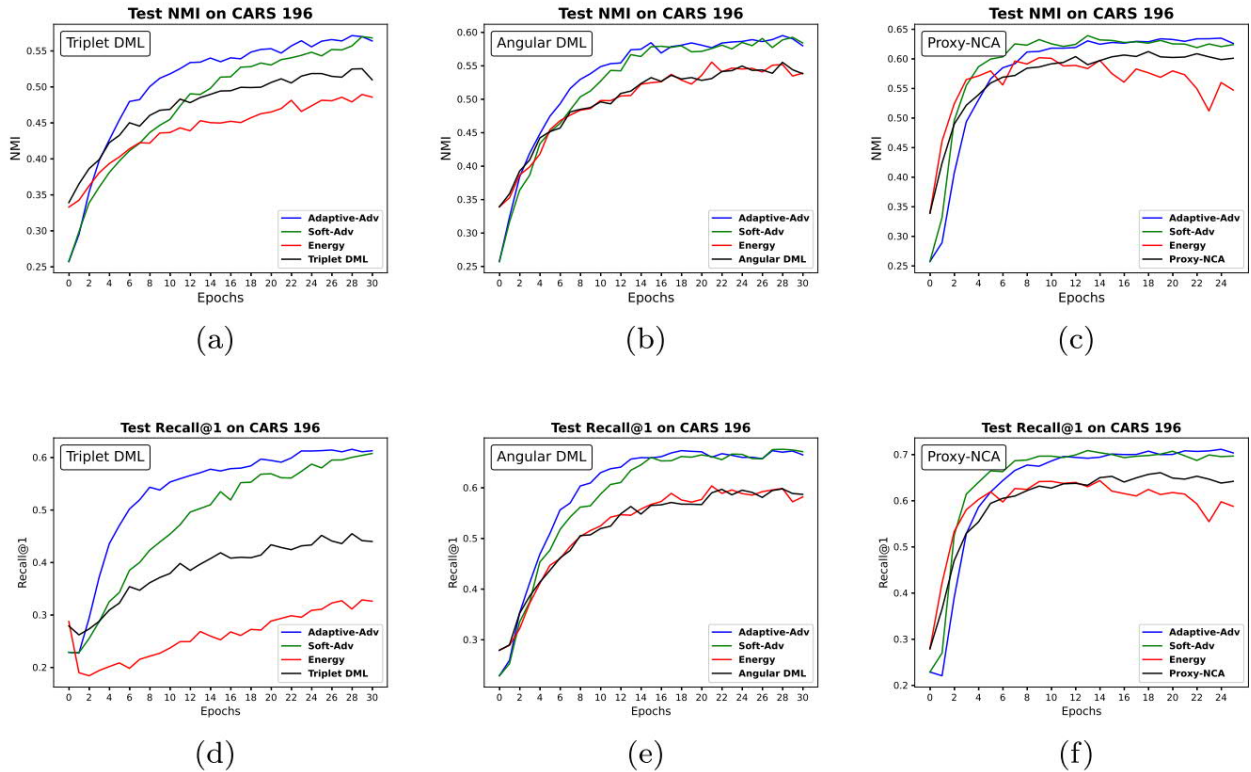

**Figure 13- NMI and Recall@1 of the evaluated methods on the CARS-196 dataset.**

## Ablation Study

In this experiment, we provide ablation studies of both *general discriminative feature learning* and the *class adversarial module* and investigate the contribution of each component individually. We choose Triplet loss as the baseline and derive two reduced methods of the proposed *Adapt-Adv* as follows:

1. Tri+GDFL[1]: indicates triplet DML using *general discriminative feature vector*. Here, we omit the class adversarial module.
2. Tri+ Adv: shows triplet DML along with the *class adversarial module*. Here, we discard the *general discriminative features* and use the output of the last hidden layer as the feature vector.

The following table reports the results. As the results show, we can derive the following conclusions:

i. Both *general discriminative feature learning* and the *class adversarial module* are effective and improve the performance of the baseline DML

ii. The *general discriminative feature learning* has a more important role to improve the baseline method.

**Table 6- Ablation studies of general discriminative feature learning and the class adversarial module on the Oxford 102 Flowers dataset.**

| Method | Extension | NMI | Recall@1 | Recall@2 | Recall@4 | Recall@8 | kNN-Acc |
|---|---|---|---|---|---|---|---|
| | Base | 70.16 | 81.55 | 88.89 | 93.59 | 96.58 | 82.22 |
| Triplet Loss: | Tri+GDFL | 73.66 | 86.41 | 91.86 | 95.09 | 97.6 | 88.12 |
| | Tri+ Adv | 72.73 | 84.54 | 90.68 | 94.71 | 97.36 | 86.03 |
| | Adapt-Adv | 76.59 | 89.32 | 94.26 | 96.48 | 97.84 | 90.39 |

## Hyper Parameters Analysis

We evaluate the effect of the adversarial coefficient $\lambda$ in this experiment. To this end, we select the Triplet loss as the baseline DML and train the Adapt-Adv DML on the Oxford 102 Flowers dataset at different $\lambda$ values. The ZSL setting as specified in Table 1 is adopted in the experiment. Figure 14 depicts the NMI and Recall@1 of the Adapt-Adv DML vs. $\lambda$ values on unseen classes.

---

[1] General Discriminative Feature Learning

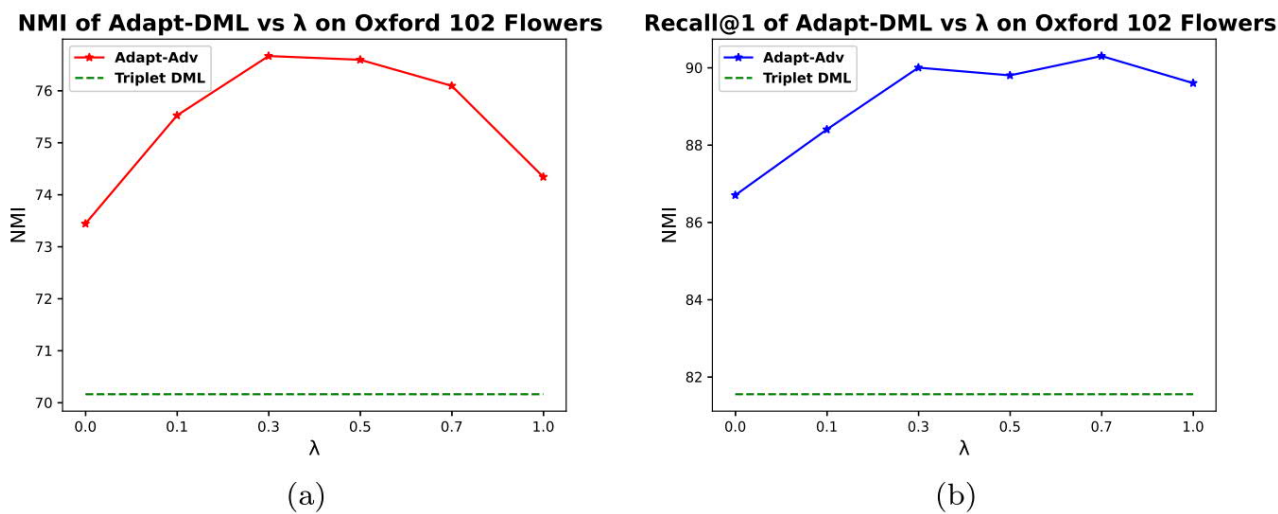

**Figure 14- NMI and Recall@1 of Adapt-Adv DML on the Oxford 102 Flowers dataset vs. λ values.**

The results indicate the effectiveness of the class adversarial module. by choosing $\lambda = 0$, our method reduces to Triplet DML with only *general discriminative feature learning* and the performance is unsatisfactory in comparison with the other $\lambda$ values. As $\lambda$ increases, the quality of the learned embedding peaked around the range (0.3, 0.7) and our method surpasses the baseline by a large margin.

### Visualization of Attended Feature Maps

In the next experiment, we illustrate the attended feature maps of the proposed feature extractor in the CUB-200-2011 dataset. Figure 15 shows some pre-process training images along with attended feature maps of selected layers. As seen, the attended features cover the most important parts of the images and each layer attends to different regions of the input images. The *inception_4d* feature maps capture more general and visual patterns of the images whereas *inception_4e* and inception_5a focused on more discriminative parts of the birds. Thus, the concatenated feature vector contains weighted useful patterns of the input image that is useful to discriminate both seen and unseen classes during the test stage.

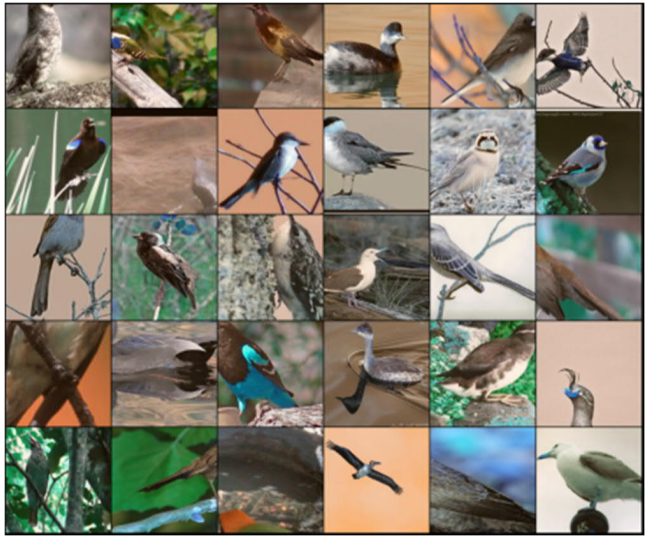
(a) Some transformed training images

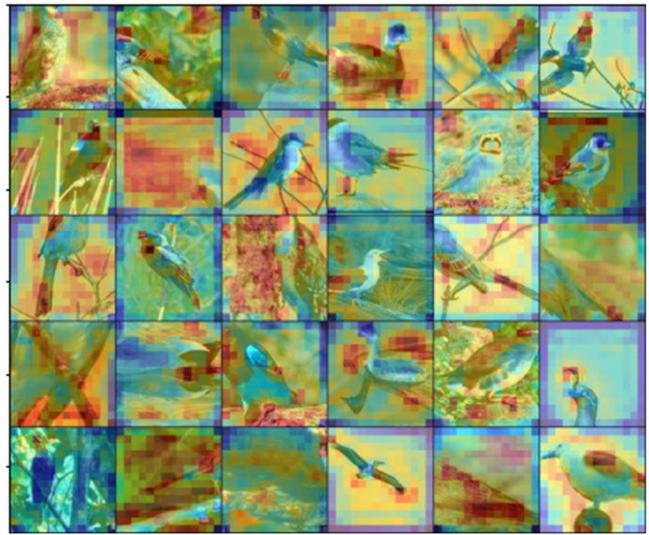
(b) Attended Feature maps of *inception_4d* layer

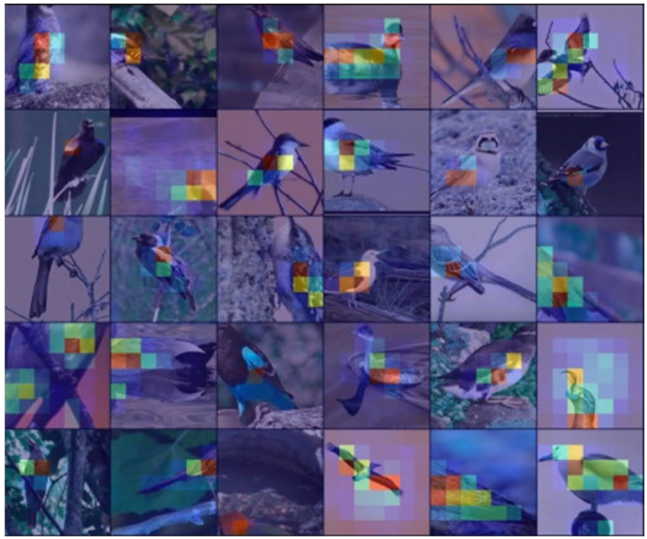
(c) Attended Feature maps of *inception_4e* layer

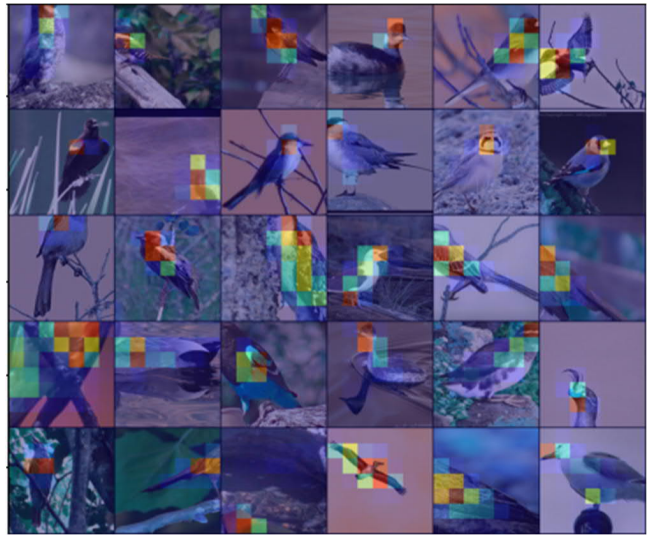
(d) Attended Feature maps of *inception_5a* layer

**Figure 15- Attended Feature maps of some training images in the CUB-200-2011 dataset.**

## 6. Conclusion and Future Work

This paper presents a novel framework to enhance the generalization of DML methods in a ZSL setting. To this end, we develop a generalized and discriminative feature learning approach and also utilize an adaptive (soft) class adversarial module. Our work can be applied to many baseline DML algorithms and improves their performance on unseen classes by a large margin. The proposed methods learn a general representation covering the most important discriminative parts of the input image that is not only specific to the observed categories.

We evaluate the proposed methods on some challenging datasets in machine vision domains in a ZSL setting. The obtained results are very encouraging, and the proposed methods consistently outperform the baselines on all evaluated datasets. It confirms the necessity and significance of our idea that adopting some *general yet discriminative feature learning*, as well as *an adaptive confusion mechanism*, is indeed helpful for most DML applications.

In future work, we intend to examine our feature learning approach on different popular deep neural network architectures and utilize the proposed framework in other applications of DML. Besides, we aim to extend the work for semi-supervised learning.

## Acknowledgment

We would like to acknowledge the Machine Learning Lab in the Engineering Faculty of FUM for their kind and technical support.